\documentclass[10pt,twocolumn,letterpaper]{article}

\usepackage{wacv}              

\usepackage{graphicx}
\usepackage{amsmath}
\usepackage{amssymb}
\usepackage{booktabs}

\usepackage{placeins}

\usepackage[pagebackref,breaklinks,colorlinks]{hyperref}

\usepackage[capitalize]{cleveref}
\crefname{section}{Sec.}{Secs.}
\Crefname{section}{Section}{Sections}
\Crefname{table}{Table}{Tables}
\crefname{table}{Tab.}{Tabs.}

\def\wacvPaperID{*****} 
\def\confName{WACV}
\def\confYear{2025}

\begin{document}

\title{Unconstrained Large-scale 3D Reconstruction and Rendering across Altitudes}

\author{Neil Joshi$^{1}$, Joshua Carney$^1$, Nathanael Kuo$^1$, Homer Li$^1$, Cheng Peng$^2$, Myron Brown$^1$ \\ 
$^1$The Johns Hopkins University Applied Physics Laboratory \\ 
$^2$Department of Computer Science, The Johns Hopkins University \\
{\small \tt \{{neil.joshi},{joshua.carney},{nathanael.kuo},{homer.li},{myron.brown}\}@jhuapl.edu}, \small \tt {cpeng26@jhu.edu} }

\maketitle

\begin{abstract}
Production of photorealistic, navigable 3D site models requires a large volume of carefully collected images that are often unavailable to first responders for disaster relief or law enforcement. Real-world challenges include limited numbers of images, heterogeneous unposed cameras, inconsistent lighting, and extreme viewpoint differences for images collected from varying altitudes. To promote research aimed at addressing these challenges, we have developed the first public benchmark dataset for 3D reconstruction and novel view synthesis based on multiple calibrated ground-level, security-level, and airborne cameras. We present datasets that pose real-world challenges, independently evaluate calibration of unposed cameras and quality of novel rendered views, demonstrate baseline performance using recent state-of-practice methods, and identify challenges for further research.
\end{abstract}

\section{Introduction}
\label{sec:introduction}

\begin{figure}[ht]
  \centering
   \includegraphics[width=0.95\linewidth]{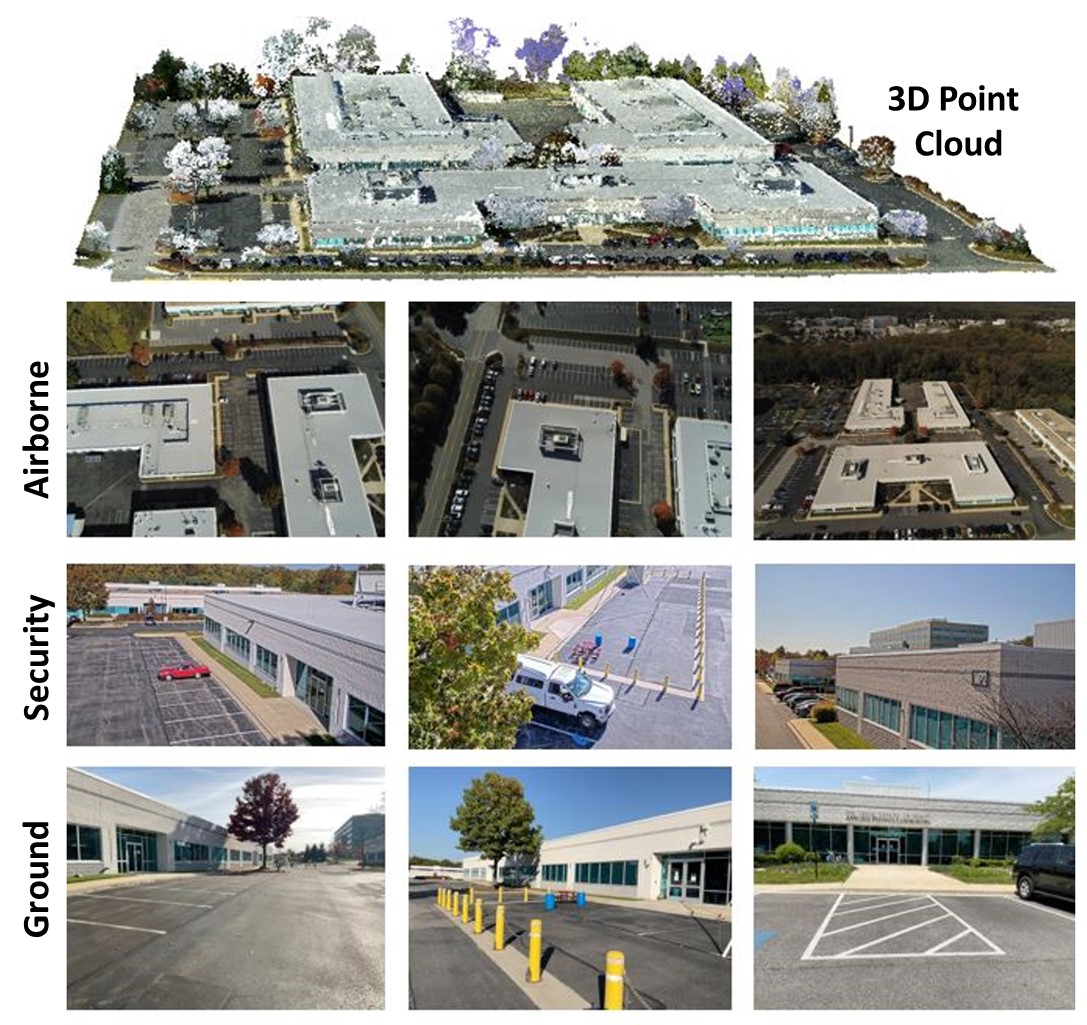}

   \caption{Ground, security, and airborne images are shown for our development benchmark dataset site, illustrating differences in viewpoint and appearance. A 3D point cloud of the full site is shown for context.}
   \label{fig:multiple-altitudes}
\end{figure}

Three-dimensional (3D) reconstruction of urban scene geometry from large numbers of satellite images \cite{Leotta_2019_CVPR_Workshops, 8014932}, airborne images \cite{Verykokou20183DRO}, or ground-level images \cite{5459148, li2018megadepthlearningsingleviewdepth} is a long-standing research problem in computer vision and graphics, with applications including urban planning, navigation, and emergency response, among many others. Methods that also produce either textured 3D models or novel rendered views of a scene enable immersive applications such as first responder training and military mission rehearsal \cite{10.3389/frvir.2021.645153}. For recent reviews of classical 3D reconstruction and modern view synthesis methods, see \cite{tewari2022advancesneuralrendering, s24072314}. These methods typically require a large volume of carefully collected images that are unavailable to first responders for disaster relief or law enforcement. In practice, the available images are often sparse and disparate in viewpoints, acquisition time, and camera types. These conditions are currently under-explored in the literature due to a lack of benchmark datasets with curated ground truth. 

In this work, we propose a benchmark dataset for camera calibration, 3D reconstruction, and novel view synthesis that emphasizes small numbers of images from ground-level cameras, security cameras, and airborne cameras that have observed a scene, as shown in Figure \ref{fig:multiple-altitudes}. Challenges in this real-world setting include limited numbers of input images from different times, heterogeneous unposed cameras, and extreme viewpoint differences for images collected from varying altitudes and at different scales. We leverage large data collection efforts to ensure that ground truth is properly measured by first acquiring a dense collection of the scene with accurate devices. The collected data is then split into various subsets to pinpoint specific challenges in camera calibration and scene reconstruction.

Prior work that motivates our approach is reviewed in Section \ref{sec:prior-work}. In Sections \ref{ssec:source-data} and \ref{sssec:methods-curation}, we describe the images and metadata contributing to our benchmark and processes for data curation. Our benchmark includes development datasets with reference values for self-evaluation and test datasets with sequestered reference values for independent evaluation \cite{ultrra-leaderboard}. Baseline algorithms for demonstrating challenges and exploring state-of-practice performance are presented in Section \ref{sssec:methods-baselines}. Our baselines build on well-supported open source software to encourage broad public experimentation. In Section \ref{sssec:methods-metrics}, we propose a metric evaluation approach that is suitable for evaluating camera pose and image similarity in real-world settings where environmental factors cannot be fully controlled. Experimental results are presented in Section \ref{sec:experiments} and conclusions in Section \ref{sec:conclusion}. Specific contributions of this work include:
\begin{itemize}
\item We present, to our knowledge, the first public benchmark dataset combining multiple ground-level, security, and airborne cameras of outdoor scenes to support research in camera calibration and view synthesis, particularly capturing challenges related to image sparsity, multiple camera types, multiple altitudes, and varying date and time.
\item We propose a methodology for evaluating camera calibration and view synthesis methods in real-world settings, demonstrate utility with our public dataset and state-of-practice baseline algorithms, and identify limiting factors for further research.
\end{itemize}

\section{Prior Work}
\label{sec:prior-work}

\textbf{Camera calibration:} Structure-from-Motion (SfM) pipelines for camera calibration with non-sequential images include Bundler \cite{Snavely2008ModelingTW}, COLMAP \cite{schoenberger2016sfm}, MVE \cite{Fuhrmann2014MVEA}, OpenMVG \cite{moulon2016openmvg}, Theia \cite{theia-manual}, and GLOMAP \cite{pan2024glomap}. Our camera calibration baseline leverages the COLMAP framework \cite{schoenberger2016sfm} due to its significant influence within the research community, integration with Nerfstudio \cite{Tancik_2023} for novel view synthesis, and ease of incorporating new algorithms.

\textbf{Novel view synthesis:} Classical pipelines for textured 3D surface reconstruction, such as OpenMVS \cite{openmvs2020} and Meshroom \cite{alicevision2021}, emphasize reconstruction of accurate 3D mesh geometry followed by blended texture mapping of input images. While this enables rendering of novel views, even small errors in reconstructed 3D geometry can lead to jarring visual imperfections in rendered images. By contrast, novel view synthesis methods aim to more directly render images of a scene from novel viewpoints based on a limited set of input images, optimizing for rendered image quality and sometimes also 3D geometry. These methods have received significant attention in recent years due to the enormous successes of neural radiance field (NeRF) \cite{Mildenhall20eccv_nerf} and 3D Gaussian Splatting (3DGS) \cite{kerbl3Dgaussians} representations. Tancik et al. \cite{Tancik_2023} recently proposed the modular Nerfstudio software framework to promote community-driven development of novel view synthesis research. Our view synthesis baseline leverages 3DGS implementated in Nerfstudio to simplify exploration of new methods as they are implemented in that framework.

\textbf{Image similarity evaluation:} The most commonly reported metrics for novel view synthesis are the structural similarity index measure (SSIM) \cite{ssim2004}, peak signal-to-noise ratio (PSNR), and learned perceptual image patch similarity (LPIPS) \cite{zhang2018perceptual}. The limitations of these low-level pixel or patch-based measures have been widely reported \cite{nilsson2020understandingssim, martin2024nerfviewsynthesissubjective, cheating}. DreamSim \cite{fu2023dreamsim, sundaram2024doesperceptualalignmentbenefit} is a recently proposed learned metric for perceptual image similarity that captures mid-level similarities in image layout, object pose, and semantic content. DreamSim is effective in capturing perceptual similarity as judged by humans and robustly identifies object similarity across poses and lighting changes. In real-world settings, small image variations such as differences in lighting, blur, and parallax are impractical to control. We use DreamSim for robust evaluation of real-world rendered image quality.

\textbf{Public datasets:} Publicly available benchmark datasets are important for enabling reproducible research and for evaluating new ideas in context with prior work. Well-calibrated datasets with images of real-world outdoor scenes are available for ground-level collection ~\cite{phototourism2006, 10.1145/3072959.3073599, Yao2019BlendedMVSAL, schoeps2017cvpr, 5995626, Tancik2022BlockNeRFSL, isprs-archives-XLVIII-1-W3-2023-219-2023} and airborne collection ~\cite{Lu_2023_ICCV, 10.1145/3072959.3073599, Yao2019BlendedMVSAL, UrbanScene3D, Turki2021MegaNeRFSC, xiangli2022bungeenerf, isprs-archives-XLVIII-1-W3-2023-219-2023}. MatrixCity \cite{li2023matrixcity} includes synthetic ground-level and aerial images at city scale. The ISPRS benchmark for multi-platform photogrammetry includes images of buildings collected jointly with ground-level and airborne images\cite{isprs-annals-II-3-W4-135-2015}. Our dataset includes images from ground, security, and airborne altitudes to enable research in cross-view camera calibration and view synthesis.

\section{Source Data}
\label{ssec:source-data}

Our work leverages data collected by a large group of engineers and scientists at the Johns Hopkins University Applied Physics Laboratory and the Massachusetts Institute of Technology Lincoln Laboratory\cite{cjk5-gf33-24}. Images were collected using a variety of mobile phones and other ground-level cameras, security cameras, and airborne drone cameras. Many of the cameras were equipped with Global Positioning System (GPS) receivers with Real-Time Kinematic (RTK) positioning capability to enable camera location accuracies of 1-5cm. Ground Control Points (GCPs) were surveyed for each site using RTK GPS.

Each camera was calibrated using commercial photogrammetry software, leveraging SfM \cite{ullman1979interpretation} and constrained by RTK GPS coordinates for either the camera locations or GCP locations selected manually in a subset of images. For each image, a sidecar metadata file captures intrinsic and extrinsic camera parameters, local date and time, and manual annotations for transient objects and imaging artifacts.

\section{Methods}
\label{sec:methods}

\subsection{Challenge dataset curation}
\label{sssec:methods-curation}

\begin{figure}[ht]
  \centering
  \includegraphics[width=0.95\linewidth]{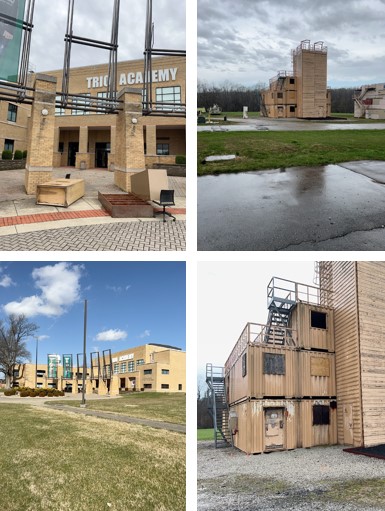}
  \caption{Images from the test datasets from Muscatatuck demonstrate collection at different times of day and year, with varying weather. View synthesis methods must model time-dependent appearance variations to produce accurate rendered images.}
  \label{fig:appearance-differences}
\end{figure}

Data released for the present public benchmark \cite{2zs6-ht63-24} include images collected at an office park in Maryland, shown in Figure \ref{fig:multiple-altitudes} and at the Muscatatuck Training Center in Indiana. Images were collected at multiple times of day and year. Figure \ref{fig:appearance-differences} illustrates time-dependent appearance differences that must be modeled in view synthesis to produce accurate rendered images for specified timestamps. 

\begin{figure*}[ht]
  \centering
  \includegraphics[width=\textwidth]{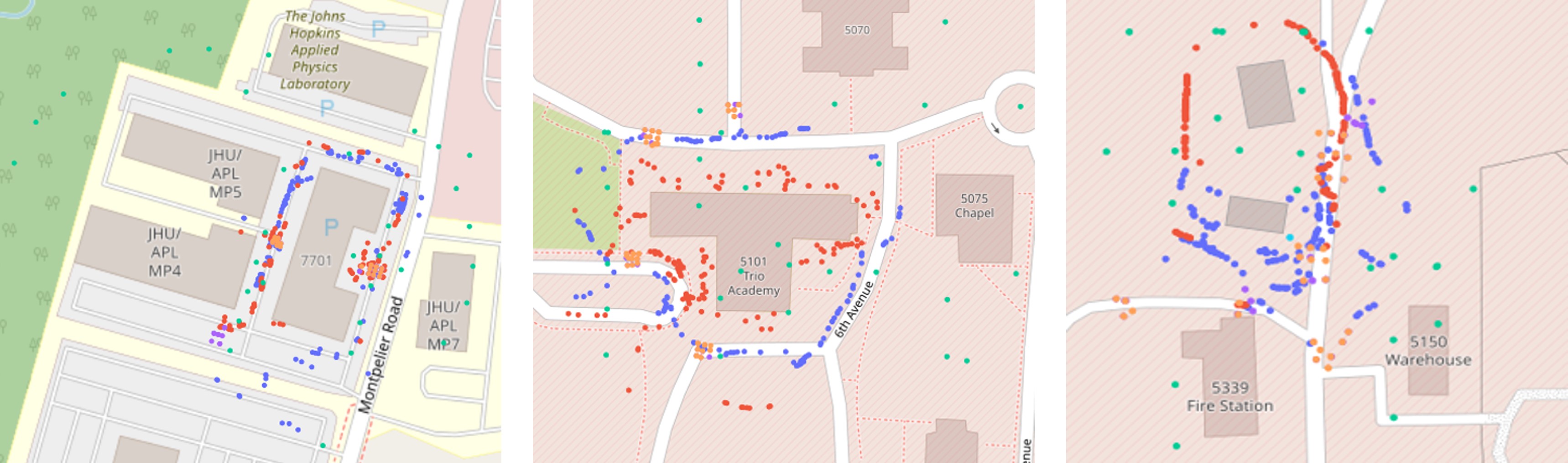}
  \caption{Approximate camera locations for the base challenge datasets are illustrated on a map for the development site at Laurel, Maryland (left) and the test sites at Muscatatuck, Indiana (center and right). Cameras shown green are airborne, red and blue are ground, and others are security. More challenging versions of datasets were produced by reducing image counts for each camera.}
  \label{fig:maps}
\end{figure*}

Based on the dense collection and ground truth position measurements from devices, challenge datasets are produced to explore camera calibration and view synthesis performance in a broad range of real-world settings. Particularly, we propose four challenges in increasing complexity order: single camera, multiple cameras, varying altitudes, and reconstructed area. For all challenges, a limited number of images from each camera is provided. Approximate camera locations for base challenge datasets are shown in Figure \ref{fig:maps}. More difficult datasets were produced for each challenge by reducing image counts from each camera based on input image DreamSim scores compared to reference images.

\textbf{Single camera:} Images were collected with a single ground-level perspective camera focused on a small area of the scene. We note that images are taken at different times, which introduces complexities due to inconsistent appearances. Furthermore, these images often contain transient objects such as cars and people. As such, it is challenging to produce a canonical and photorealistic 3D reconstruction due to shape-radiance ambiguity~\cite{zhang2020nerf++,martinbrualla2020nerfw}. 

\textbf{Multiple cameras:} Images were collected with the same single camera, plus images from additional ground-level and security-level cameras, collected at varying times of day and year, and focused on the same small area of the scene. Multiple camera types often lead to suboptimal calibration accuracy due to the need to estimate more complex intrinsics; furthermore, security cameras are stationary, which can lead to ill-defined SfM formulation.

\textbf{Varying altitudes:} Images were collected from multiple ground-level and security cameras focused on the same small area of the scene, plus images from airborne cameras with much different viewpoints. Calibration becomes difficult in this challenge due to the vastly different perspectives. Similarly, photorealistic reconstruction suffers from significant floaters under the varying altitude scenario~\cite{li2023matrixcity}. 

\textbf{Reconstructed area:} Images were collected with multiple cameras and at varying altitudes, plus more images from each camera focused on a larger area of the scene. On top of previous challenges, large-scale camera calibration often suffers from visual ambiguities, where images taken at very different locations can have repeating structures such as windows and doors, leading to erroneous calibration. These are often referred to as doppelgangers~\cite{cai2023doppelgangers, xiangli2024doppelgangersimprovedvisualdisambiguation}. 

\begin{figure}[ht]
  \centering
  \includegraphics[width=0.9\linewidth]{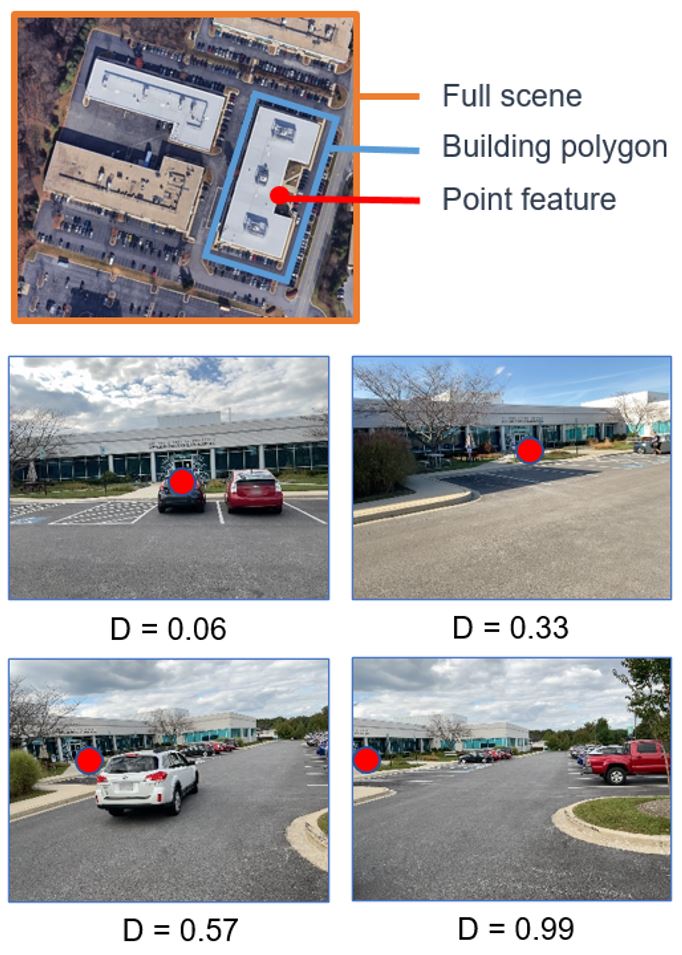}
  \caption{Challenge datasets are produced by defining world coordinates for points or polygons in a scene (top) and then sampling images that observe those points based on normalized distance ($D \in [0,1]$) to image center (bottom).}
  \label{fig:wrivacraft}
\end{figure}

For each test site, we identify world coordinates of points or polygons identifying features of interest in those scenes, as shown in Figure \ref{fig:wrivacraft}. We then project those world coordinates into well-calibrated and geolocated cameras to identify images that observe those features. A digital surface model derived from either lidar or photogrammetry is employed to reject images with static scene occlusions such as buildings or trees between the camera and the selected world coordinates. Images are sampled based on camera type, camera altitude, normalized distance of projected world coordinates to image center, presence of imaging artifacts, presence of transient objects, time of day, season, and other factors.

For each challenge dataset, we separately evaluate camera calibration for unposed input images and novel view synthesis for input images with known camera locations. End-to-end view synthesis performance given unposed cameras can also be evaluated using our datasets, though we do not emphasize this for leaderboard evaluation or in our experimental results (Section \ref{sec:experiments}).

\subsection{Baseline algorithms}
\label{sssec:methods-baselines}

\textbf{Camera calibration:} Our camera calibration baseline solution leverages the COLMAP framework \cite{schoenberger2016sfm} for SfM\cite{ullman1979interpretation}. For feature point extraction and matching, we employ SuperPoint \cite{8575521} and LightGlue \cite{lindenberger2023lightglue} from the Hierarchical Localization (hloc) toolbox\cite{hloc2019}. We estimate intrinsic camera parameters independently for each input image to accommodate multiple camera models. Since input camera locations are not provided for the camera calibration challenge, COLMAP produces camera pose predictions in a local Cartesian coordinate system or multiple coordinate systems if all images cannot be successfully aligned together. In this case, the largest group of images successfully aligned is used to determine a single local coordinate system. In evaluation, local coordinates are aligned with a known reference coordinate system using Procrustes analysis\cite{10.5555/59560} to produce error metrics.

\textbf{Novel view synthesis:} Our view synthesis baseline uses SplatFacto, an implementation of 3D Gaussian Splatting \cite{kerbl3Dgaussians} in Nerfstudio\cite{Tancik_2023}. Notably, SplatFacto uses gsplat as its Gaussian rasterization back-end \cite{ye2024gsplatopensourcelibrarygaussian}. We used gsplat version 1.4, which demonstrates significant performance improvements over previous versions. For the view synthesis challenge, input camera locations are provided. We align the local coordinates from camera calibration, described above, to these reference camera locations to enable rendering of images from our model with camera parameters provided in the reference world coordinate system. We also modify the camera calibration pipeline described above to better resolve the issue of multiple local coordinate systems by independently applying Procrustes analysis to each.

\begin{figure*}[ht] 
  \includegraphics[width=\textwidth]{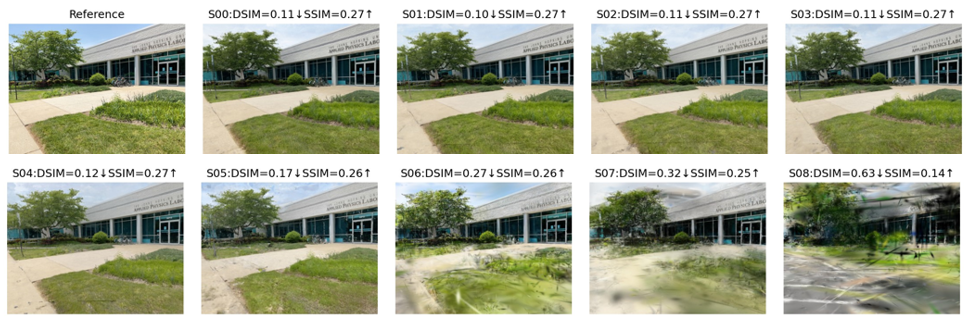}
  \caption{DreamSim (labeled DSIM, lower is better) and SSIM (higher is better) are shown for a single reference image compared to novel views rendered from a sequence of 3DGS models. At each step in the sequence, the number of input images for training is reduced in the order of $\{150, 125, 100, 75, 50, 25, 15, 10, 5\}$. DreamSim scores better capture the range of visual similarity.}
  \label{fig:dreamsim-ssim-comparison}
\end{figure*}

\subsection{Metric evaluation}
\label{sssec:methods-metrics}

To evaluate camera calibration for unposed input images, we compute relative geolocation error for each input camera and report the 90th percentile spherical error (SE90) as our summary metric. Percentile statistics were selected because geolocation error is unbounded and subject to severe outliers. The 90th percentile was selected to encourage accurate calibration for images from all cameras. No information is provided to identify world coordinates, so camera pose predictions are expected in a local Cartesian coordinate system. Local coordinates are aligned to reference world coordinates using Procrustes analysis \cite{10.5555/59560}. Images identified in the pose estimation algorithm to be poorly calibrated are not included in the fit to increase reliability, but they are included in metric evaluation.

To evaluate novel view synthesis, we map reference camera projections to local coordinates to request rendered images. We compute the DreamSim mid-level perceptual similarity metric \cite{fu2023dreamsim} between rendered images and held-out reference images to assess image similarity. DreamSim has been shown to be effective in capturing perceptual similarity as judged by humans. In our experience, none of the commonly reported low-level metrics produce reasonable relative rankings of image similarity in real-world settings, due to severe sensitivity to often visually imperceptible appearance variations, as discussed in Section \ref{sec:prior-work}. For a practical example comparing DreamSim and SSIM with real-world images, see Figure \ref{fig:dreamsim-ssim-comparison}. Our summary metric for each dataset is the mean of DreamSim scores for all rendered images.

\begin{figure}[ht!]
  \centering
  \includegraphics[width=0.9\linewidth]{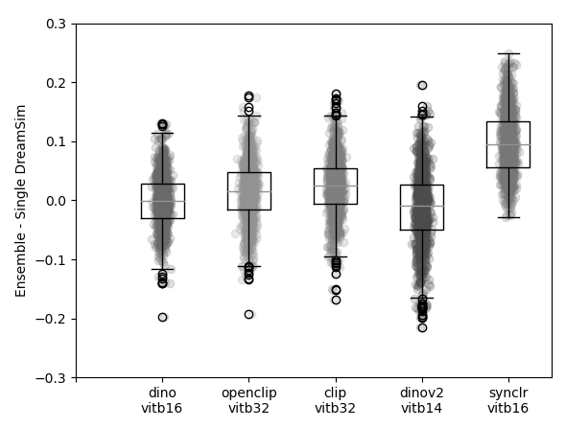}
  \caption{We use the OpenCLIP single-branch variant of DreamSim for leaderboard evaluation to minimize file sizes, memory requirements, and run times. Differences between ensemble scores and individual model scores for all pairs of our development dataset images are shown.}
  \label{fig:dreamsim-model-comparison}
\end{figure}

DreamSim was trained by concatenating multiple large vision model feature embeddings and fine-tuning on human perceptual judgements. The ensemble model is large and computationally expensive, so for leaderboard evaluation with limited resources, we report DreamSim scores produced using the OpenCLIP single-branch variant, which produces reasonably similar results (Figure \ref{fig:dreamsim-model-comparison}). Since new model weights can be released with new software versions, we recommend DreamSim 0.2.1 to ensure reproducibility of our results.

Limitations of the DreamSim metric are acknowledged in \cite{fu2023dreamsim}, such as inherited bias from the pre-trained vision model backbones and significant emphasis on foreground objects and semantic content, which leads to less sensitivity to background details or contextual elements. We have observed a few clear examples of these issues, so we are careful to select reference images with obvious foreground objects for evaluation to mitigate this.

\begin{figure*}[ht]
  \centering
  \includegraphics[width=\textwidth]{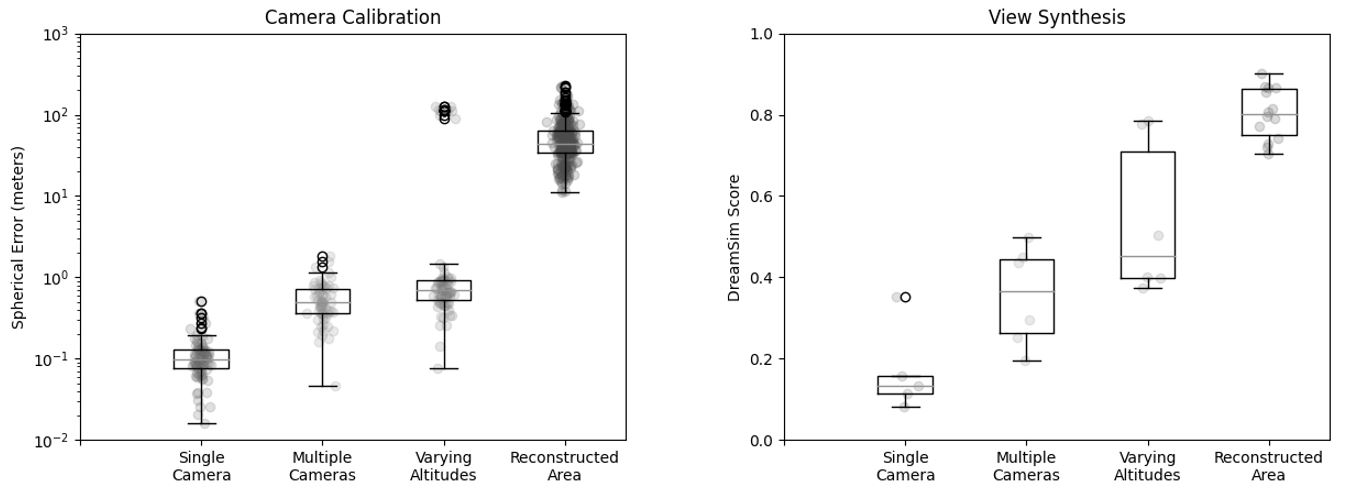}
  \caption{Development dataset baseline results are shown with all input images. Camera geolocation errors and DreamSim scores (lower is better) increase with complexity of the four challenges: single camera model, multiple camera models, varying altitudes, and increased reconstructed area.}
  \label{fig:dev-plots}
\end{figure*}

\section{Baseline Results and Challenges}
\label{sec:experiments}

Our experiments provide a check on dataset quality and fairness, to establish minimum expectations for performance using state of practice algorithms, and to highlight challenges that deserve more attention. We apply our COLMAP and Nerfstudio baseline algorithms (Section \ref{sssec:methods-baselines}) to the development and test challenge datasets (Section \ref{sssec:methods-curation}) and evaluate using metrics described in Section \ref{sssec:methods-metrics}. 

\begin{figure}[ht]
  \centering
  \includegraphics[width=0.9\linewidth]{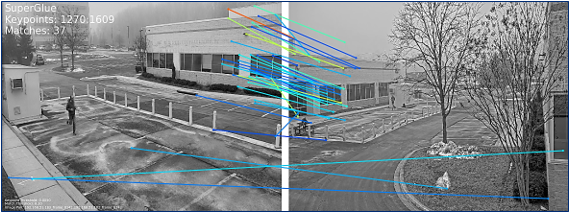}
  \caption{Feature matching challenges for large-scale scenes include cross-view appearance differences and visually repetitive features, both demonstrated here. Feature matching with SuperGlue \cite{9157489} fails for this pair of security camera images taken from opposing viewpoints.}
  \label{fig:cross-view-is-hard}
\end{figure}

\begin{figure}[ht]
  \centering
  \includegraphics[width=0.9\linewidth]{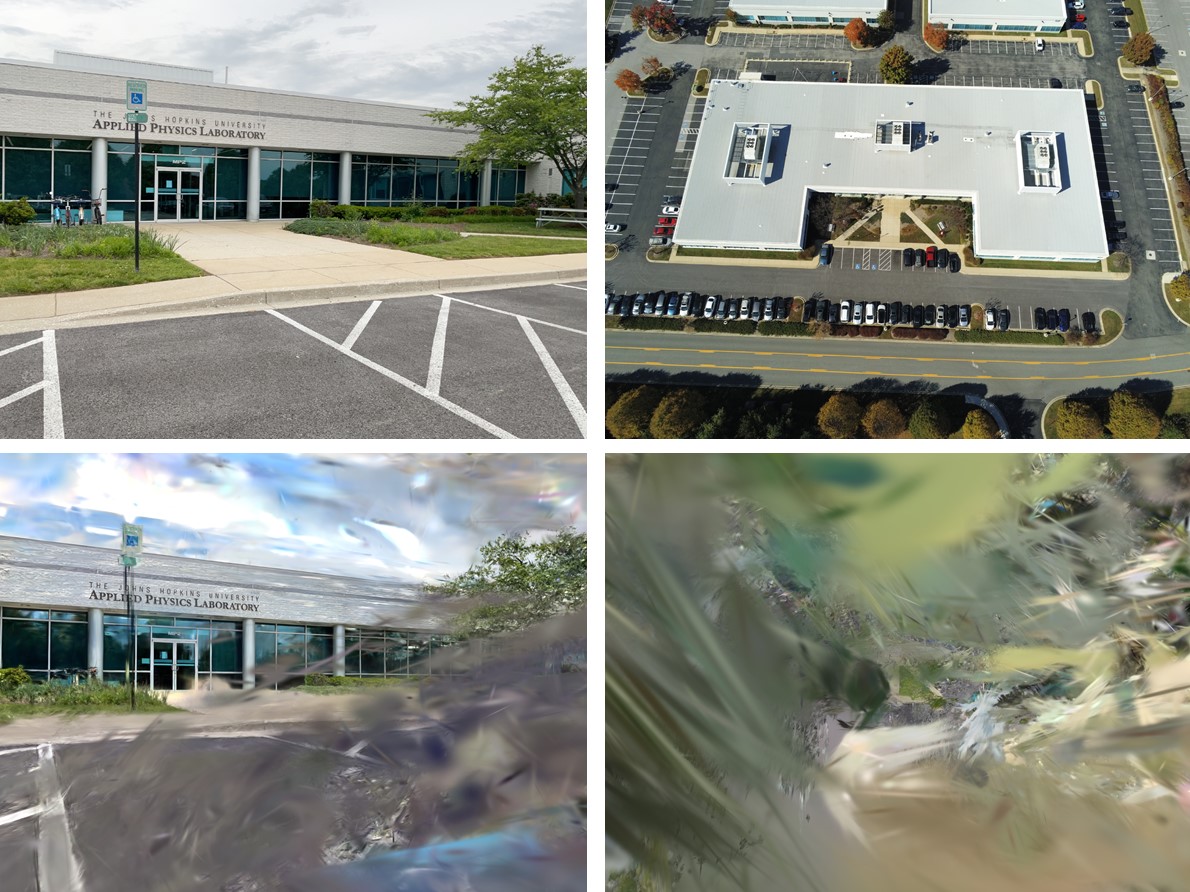}
  \caption{Incorrectly predicted geometry not directly observed by input images can occlude well-modeled portions of a scene when rendered from a novel viewpoint (example reference image shown left top and render left bottom). Similarly, incorrectly predicted geometry in the sky from ground-level views can occlude images rendered from airborne viewpoints (example shown right).}
  \label{fig:rendering-artifacts}
\end{figure}

Baseline leaderboard results for camera location (SE90) and image similarity (mean DreamSim) scores are summarized for the development and test sets in Table \ref{tab:leaderboard-scores}. Box and scatter plots in Figure \ref{fig:dev-plots} show camera location and view synthesis scores reported for each evaluated image, better illustrating the range of performance. While our baselines perform reasonably well for the single and multiple camera challenges, they perform poorly for varying altitude and reconstructed area challenges. Observed limitations suggest research challenges to be explored with our benchmark datasets:

\begin{itemize}
\item \textbf{Cross-view camera calibration:} Our varying altitudes and reconstructed area challenges include cameras from both ground-level and airborne perspectives. Matching features between pairs of images with extreme viewpoint differences is very challenging, as illustrated in Figure \ref{fig:cross-view-is-hard}. This is especially challenging for scenes with visually repetitive features. Methods for cross-view matching with ground and airborne images include ~\cite{7035866, ZHU2024311, chen2024geometryawarefeaturematchinglargescale}.
\item \textbf{Doppelgangers:} Our reconstructed area challenges include examples of so-called doppelgangers, or visually similar images that depict different parts of a scene. Methods to identify these ambiguous image pairs include ~\cite{cai2023doppelgangers, xiangli2024doppelgangersimprovedvisualdisambiguation}.
\item \textbf{Inaccurate occluding geometry in novel view synthesis:} Portions of the scene not sufficiently observed by input images may be inaccurately modeled, resulting in incorrect geometry that occludes well-modeled portions of the scene when rendered from novel views. Examples are shown in Figure \ref{fig:rendering-artifacts}. Depth and semantic priors have been proposed to discourage these artifacts in optimization\cite{WarburgWTHK23, Wu2023ReconFusion3R, xu2024depthsplat}. Care must also be taken when combining ground-level and airborne camera viewpoints. Incorrectly modeled sky geometry from ground views can occlude well-modeled portions of the scene when rendered from airborne views.
\item \textbf{Temporally varying appearance:} Our datasets include images collected with varying times of day and season, each with visually distinct appearance, as illustrated in Figure \ref{fig:appearance-differences}. If not explicitly modeled, these variations can result in poor rendered image quality. Date and time stamps are provided as inputs to novel view synthesis challenges. Methods for modeling these variations include \cite{martinbrualla2020nerfw, xu2024splatfactownerfstudioimplementationgaussian}.
\end{itemize}

\begin{table}[ht]
  \centering
  {\small{
  \begin{tabular}{llcc}
    \toprule
    Phase & Challenge Dataset & SE90 (m) $\downarrow$ & DreamSim $\downarrow$ \\
    \midrule
    Dev & Single Camera & 0.19 & 0.17 \\
    Dev & Multiple Cameras & 0.87 & 0.35 \\
    Dev & Varying Altitudes & 108.85 & 0.54 \\
    Dev & Reconstructed Area & 87.80 & 0.80 \\
    \midrule
    Test & Single Camera & 0.10 & 0.25 \\
    Test & Multiple Cameras & 1.28 & 0.27 \\
    Test & Varying Altitudes & 94.82 & 0.71 \\
    Test & Reconstructed Area & 65.17 & 0.64 \\
    \bottomrule
  \end{tabular}
  }}
  \caption{Baseline leaderboard scores are shown for the development and sequestered test datasets. Lower is better for both scores.}
  \label{tab:leaderboard-scores}
\end{table}

\section{Conclusion}
\label{sec:conclusion}

We have presented a public benchmark dataset and leaderboard metric evaluation methodology to encourage research in camera calibration, 3D reconstruction, and novel view synthesis for challenging real-world settings, combining images from ground-level, security, and airborne cameras. Public data is available at \cite{2zs6-ht63-24}, public leaderboard at \cite{ultrra-leaderboard}, and baseline and metric implementations at \cite{ultrra-baseline}.

The datasets we have crafted emphasize a range of challenges in calibration and 3D reconstruction with multiple camera models, varying altitudes, and spatial scale. Our baseline methods build on the open source COLMAP and Nerfstudio frameworks to enable straightforward algorithm integration, experimentation, and metric evaluation to advance the state of the art for these challenging settings.

Our data curation framework can also be applied to explore a broader range of performance factors. For future work, we plan to publicly release a more comprehensive set of challenge datasets. We also plan to publicly release our data curation source code along with a large corpus of source data to allow researchers to construct their own datasets.

\section*{Acknowledgments}
This work was supported by the Intelligence Advanced Research Projects Activity (IARPA) contract numbers 2020-20081800401 and 140D0423C0076. Disclaimer: The views and conclusions contained herein are those of the authors and should not be interpreted as necessarily representing the official policies or endorsements, either expressed or implied, of IARPA or the U.S. Government.

\FloatBarrier

{\small
\bibliographystyle{ieee_fullname}
\bibliography{egbib}
}

\end{document}